\documentclass[letter, 10 pt, conference]{ieeeconf}  % Comment this line out
\IEEEoverridecommandlockouts                              % This command is only
\usepackage{graphicx}
\usepackage{color}
\usepackage[dvipsnames]{xcolor}
\usepackage[colorlinks]{hyperref}
\usepackage{subfigure}
\usepackage{makecell}
\usepackage{multirow}
\usepackage{tabularx}
\makeatletter
\def\hlinewd#1{%
\noalign{\ifnum0=`}\fi\hrule \@height #1 %
\futurelet\reserved@a\@xhline}
\makeatother
\usepackage{caption2}
\usepackage{todonotes}
\usepackage{CJK}
\usepackage{cite}
\usepackage{amsmath}
\usepackage{amsfonts}
\usepackage{amssymb}
\usepackage{booktabs}

\usepackage{booktabs}
\usepackage{hhline}

\title{\LARGE \bf
Two-Stream RNN/CNN for Action Recognition in 3D Videos}
\author{Rui Zhao$^{1}$, Haider Ali$^{2}$, and Patrick van der Smagt$^{3}$
\thanks{$^{1}$RZ is with Siemens AG \& Ludwig Maximilian University of Munich {\tt\small $\lbrace$ruizhao@siemens.com$\rbrace$}.
$^{2}$HA is with Center  for  Imaging  Science (CIS),  Johns  Hopkins  University,  Baltimore, MD, USA {\tt\small $\lbrace$hali@jhu.edu$\rbrace$}.
$^{3}$PvdS is Head of AI Research, Data:Lab, Volkswagen Group.
Work done when RZ, HA were at Institute of Robotics and Mechatronics, German Aerospace Center (DLR), 82234 Wessling, Germany.}
}

\begin{document}

\maketitle
\thispagestyle{empty}
\pagestyle{empty}

%\graphicspath{{figures/}}
\graphicspath{{/}}

\begin{abstract}

The recognition of actions from video sequences has many applications in health monitoring, assisted living, surveillance, and smart homes.
Despite advances in sensing, in particular related to 3D video, the methodologies to process the data are still subject to research.
We demonstrate superior results by a system which combines recurrent neural networks with convolutional neural networks in a voting approach.
The gated-recurrent-unit-based neural networks are particularly well-suited to distinguish actions based on long-term information from optical tracking data; the 3D-CNNs focus more on detailed, recent information from video data. The resulting features are merged in an SVM which then classifies the movement.
In this architecture, our method improves recognition rates of state-of-the-art methods by 14\% on standard data sets.

%Improved low-cost sensors, providing 3D videos at reasonable accuracy have enabled many such technologies.
%At the same time, 

%Action recognition is used in many fields including health monitoring in hospitals, assisted living for
%smart homes and robotics surveillance for military. Compared to 2D videos, 3D videos can acquire more detailed spatial and
%temporal information from the real word. This paper presents research that concerns applying deep learning methods to large
%scale 3D human action classification. Our contribution is three-fold. First, we propose a new recurrent neural network structure
%using advanced cell structure and normalization methods. This structure not only can be trained much faster, but also achieves
%state-of-the-art performance on the currently largest 3D human action dataset with 3D skeleton data alone.
%Secondly, a 3D-CNN model is finetuned with RGB videos from the action dataset. Finally, we propose a two-stream recurrent convolutional neural network structure, which utilizes the 3D-CNN model to learn video features and the proposed RNN model to learn 3D skeleton features, then feeds all the features into a simple linear SVM model and beats current best classification accuracy by a significant margin, more than 14\%.  

\end{abstract}
\section{Introduction}

%\subsection{The paper is about}
Recognition of human activity in 3D videos has received increasing attention since 2010\cite{li2010action,Shahroudy_2016_CVPR,du2015hierarchical,liu2016spatio,zhu2016co,wang20143d,wang2015action}.
Compared to 2D videos, 3D videos provide more spatial information and could be more informative.
Action recognition with 3D videos is applied in different fields, such as health monitoring for patients, assisted living for disabled people, and robot perception and cognition.

Following this line of research, this paper proposes and applies novel deep-learning methods on what is currently the largest 3D action recognition dataset.  Our results are compared with existing best approaches and are shown to be superior.
Our proposed deep-learning methods consist mainly of three parts: a novel skeleton-based recurrent neural network structure, using a 3D-convolutional \cite{tran2015learning} neural network for RGB videos, and sketching a new two-stream fusion method to combine RNN and CNN. All methods are evaluated on the NTU RGB+D Dataset\cite{Shahroudy_2016_CVPR}. The dataset was published in 2016 and contains more than 56k action samples in four different modalities: RGB videos, depth map sequences, 3D skeletal data, and infrared videos. The dataset consists of 60 different action classes including daily, health-related, and mutual actions. In this paper, we use both the 3D skeletal data and RGB videos.  

%\subsection{literature review and explain why better}
Traditional studies on 3D action recognition use different kinds of methods \cite{li2010action,evangelidis2014skeletal,vemulapalli2014human,hu2015jointly,rahmani2016histogram,gaglio2015human,chen2013real,
ni2013rgbd,sung2012unstructured,wang2012mining,xia2012view,bloom2013dynamic,lin2012human,zhang2012privacy,oreifej2013hon4d,
koppula2013learning,negin2013decision,wei2013concurrent,munaro2013evaluation,ellis2013exploring,mansur2013inverse,yang2013rgb,
carletti2013recognition,kastaniotis2013gait,liu2015coupled,huang2014sequential,lillo2014discriminative,yu2014discriminative,
wu2015watch,hu2016exemplar,chen2015utd,cheng2012human,zhang2012viewpoint,ofli2013berkeley,amiri2013non,wei2013modeling,
wang2014cross,rahmani2014hopc,liu2015multipe,song2014body,yun2012two,hu2013efficient,wolf2014evaluation,bloom2014g3di,
van2014dyadic} to compute handcrafted features, while deep-learning approaches \cite{wang20143d,lin2015deep,wang2015action,wang2015convnets,du2015hierarchical,Shahroudy_2016_CVPR,zhu2016co,liu2016spatio} are end-to-end trainable and can be applied directly on raw data. 
Focussing on the latter, for skeleton-based activity analysis, \cite{du2015hierarchical,Shahroudy_2016_CVPR,zhu2016co,liu2016spatio} used different kinds of recurrent neural networks to acquire state-of-the-art performances on various of 3D action datasets. Du et al.\ \cite{du2015hierarchical} propose an hierarchical RNN, which is fed with manually divided five groups of the human skeleton, such as two hands, two legs, and one torso. Inspired from this, Shahroudy et al.\ \cite{Shahroudy_2016_CVPR} present a novel long short-term memory (LSTM) \cite{hochreiter1997long} cell, called part-aware LSTM, which is also fed with separated five parts of skeleton. Evolved from these two ideas, Zhu et al.\ \cite{zhu2016co} provide a novel deep RNN structure, which can automatically learn the co-occurrence, similar to grouping data into five human body parts, from skeleton data. Most recently, Liu et al.\ \cite{liu2016spatio} propose a skeleton tree traversal algorithm and a new gating mechanism to improve robustness against noise and occlusion.

However, our proposed RNN structure makes a different contribution. Our method is inspired by recent normalization technologies \cite{ioffe2015batch} and a novel recurrent neuron mechanism \cite{cho2014learning}. With these advanced deep learning technologies embedded into our RNN structure, it can be trained with 13 times fewer iterations and for each iteration consumes 20\% less computational time, compared to a normal RNN model with LSTM cells. Our contribution focuses more on making the network much easier to train, less inclined to overfitting, and deep enough to represent the data. More importantly, our proposed RNN structure outperforms all other skeleton-based methods on the largest 3D action recognition dataset.

To process RGB videos, our method is inspired by different kinds of convolutional neural networks\cite{tran2015learning,donahue2015long,simonyan2014two,ji20133d}. We use a 3D-CNN\cite{tran2015learning} model on the RGB videos of the NTU RGB+D dataset. We compare the results with proposed RNN models, and fuse their output.

To combine the RNN and CNN models, we propose two fusion structures: decision and feature fusion. The first is very simple to use, whereas the second provides a better performance. For decision fusion, we illustrate a voting method based on the confidence of the classifiers. For feature fusion, we propose a novel two-stream RNN/CNN structure, shown in Fig.~\ref{fig:net_structure}, which combines temporal and spatial features from the RNN and CNN models and boost the performance by a significant margin. Our two-stream RNN/CNN structure outperforms the current state-of-the-art method \cite{liu2016spatio} more than 14\% on both cross subject and cross view settings.

To summarize, our contributions in this paper are:
\begin{itemize}
  \item[$\bullet$] A novel RNN structure is proposed, which converges 13 times faster during training and costs 20\% less computational power at each forward pass, compared to a normal LSTM;
  \item[$\bullet$] Two fusion methods %, i.e.\ decision fusion and feature fusion,
are proposed to combine the proposed RNN structure and a 3D-CNN structure \cite{tran2015learning}. The decision fusion is easier to use, while the feature fusion has superior performance.
\end{itemize}\par

%This paper consists of two more main sections, methods in Sec.~\ref{sec:Methodology} and experiments in Sec.~\ref{sec:Experiment}. Sec.~\ref{sec:Recurrent Neural Network} introduces relevant components in our proposed RNN structure, Sec.~\ref{sec:Convolutional Neural Network} illustrates the 3D-CNN model, and Sec.~\ref{sec:Two-stream R-CNN} shows further detail about the two-stream R-CNN model. These models are later evaluated in Sec.~\ref{sec:Experiment}.  

\section{Methodology}
\label{sec:Methodology}

In this section, we first introduce the concept of recurrent neural networks and batch normalization, and then describe the proposed RNN structure: a deep bidirectional gated recurrent neural network with batch normalization and dropout. Afterwards, the applied 3D-CNN model and two-stream RNN/CNN fusion architectures, i.e., decision fusion and feature fusion, are introduced.

\subsection{Recurrent Neural Network}
\label{sec:Recurrent Neural Network}
\subsubsection{Vanilla Recurrent Neural Network}
Recurrent neural networks can handle sequence information with varied lengths of time steps. This transforms the input \begin{math} \mathbf{X} \end{math} to a internal hidden state \begin{math} \mathbf{h}_{t} \end{math} at each time step. The network passes the state \begin{math} \mathbf{h}_{t} \end{math} along with the next input \begin{math} \mathbf{x}_{t+1} \end{math} to the neuron, time step after time step. The neuron learns when to remember and forget information with nonlinear activation functions:
\begin{equation}\label{eq:rnn ht}
\mathbf{h}_{t} = \sigma \left( \mathbf{W} \left(\begin{array}{ccc} \mathbf{x}_{t} \\ \mathbf{h}_{t-1} \\ \end{array} \right) \right)
\end{equation} 
\begin{equation}\label{eq:rnn yt}
\mathbf{y}_{t} = \sigma \left( \mathbf{V}  \mathbf{h}_{t} \right)
\end{equation}
where \begin{math} t \in \{1,\dots,T \} \end{math} represents time steps, and \begin{math} \sigma \end{math} represents a nonlinear activation function such as a standard logistic sigmoid function $\mathrm{sigm}(x) = 1 / (1+\mathrm{e}^{-x})$
or a hyperbolic tangent function $\tanh(x)$.

Multiple layers of RNN can be stacked to increase the complexity:
\begin{equation}\label{eq:rnn htl}
\mathbf{h}_{t}^{(l)} = \sigma \left( \mathbf{W}^{(l)} \left(\begin{array}{ccc} \mathbf{h}_{t}^{(l-1)} \\ \mathbf{h}_{t-1}^{(l)} \\ \end{array} \right) \right)
\end{equation}
\begin{equation}\label{eq:rnn ht0}
\mathbf{h}_{t}^{(0)} := \mathbf{x}_{t}
\end{equation}
\begin{equation}\label{eq:rnn yt0}
\mathbf{y}_{t} = \sigma \left( \mathbf{V}\mathbf{h}_{t}^{(L)} \right)
\end{equation}
where \begin{math} l \in \{ 1,...,L\} \end{math} denotes the layer number.

In practice, a vanilla RNN does not remember information over a longer time; a problem which is related to the vanishing gradient problem.
%For example, an RNN use \begin{math} \tanh \end{math} as activation function. Input information passed along time steps through RNN units can be understand as \begin{math}\tanh(\tanh(\tanh(...))) \end{math}\,. During backpropagation, the partial gradient of the errors with respect to weights, \begin{math}\partial E / \partial w \end{math}, is calculated using the chain rule. Mathematically, this appears as: \begin{math}\partial E / \partial w=\tanh'\tanh'\tanh'(...) \end{math}. And \begin{math}\tanh'=1-\tanh^{2} \end{math}, so \begin{math}\tanh' \in [0,1) \end{math}, which means that the partial gradient equals a quantity multiplied frequently by an amount smaller than one, which can be extremely small, almost 0, for an instance: \begin{math} 0.9^{100}\doteq 0.0000266 \end{math}. \par
%making it very hard for an RNN to update weights with backpropagation. 

\subsubsection{Long Short-Term Memory}
This problem can be solved by LSTM \cite{hochreiter1997long} which stores information in gated cells at the neurons.
This allows errors to be backpropagated through hundreds or thousands of time steps:
\begin{equation}\label{eq:lstm gates}
\left( \begin{array}{ccc} \mathbf{i} \\ \mathbf{f} \\ \mathbf{o} \\ \hat{\mathbf{c}}_{t} \\ \end{array} \right) = \left( \begin{array}{ccc} \mathrm{sigm} \\ \mathrm{sigm} \\ \mathrm{sigm} \\ \tanh \\ \end{array} \right) \mathbf{W} \left(\begin{array}{ccc} \mathbf{x}_{t} \\ \mathbf{h}_{t-1} \\ \end{array} \right)
\end{equation}
\begin{equation}\label{eq:lstm ct}
\mathbf{c}_{t} = \mathbf{f} \odot \mathbf{c}_{t-1}+\mathbf{i}\odot \hat{\mathbf{c}_{t}}
\end{equation}
\begin{equation}\label{eq:lstm ht}
\mathbf{h}_{t} = \mathbf{o} \odot \tanh(\mathbf{c}_{t})
\end{equation}
where \begin{math}\mathbf{i},\mathbf{f}\end{math} and \begin{math} \mathbf{o} \end{math} denote input gate, forget gate, and output gate, respectively. \begin{math} \hat{\mathbf{c}}_{t} \end{math} represents new candidate values, which could be added to the cell state \begin{math} \mathbf{c}_{t} \end{math}. We use  \begin{math} \odot \end{math} for element-wise multiplication.

\subsubsection{Gated Recurrent Unit}
An improvement to LSTM called gated recurrent unit (GRU) was proposed in \cite{cho2014learning}. GRU has a simpler structure and can be computed faster. The three gates from LSTM are combined into two gates, respectively updating gate \begin{math} \mathbf{z} \end{math} and resetting gate \begin{math} \mathbf{r} \end{math} in GRU. GRU also combines cell state \begin{math} \mathbf{c}_{t}\end{math} and hidden state \begin{math} \mathbf{h}_{t}\end{math} into one state \begin{math} \mathbf{h}_{t}\end{math}. The mathematical description is as follows:
\begin{equation}\label{eq:gru gates}
\left(\begin{array}{ccc} \mathbf{z} \\ \mathbf{r} \\ \end{array} \right) = \left(\begin{array}{ccc} \mathrm{sigm} \\ \mathrm{sigm} \\ \end{array} \right) \left( \mathbf{W} \left(\begin{array}{ccc} \mathbf{x}_{t} \\ \mathbf{h}_{t-1} \\ \end{array} \right) \right)
\end{equation}  
\begin{equation}\label{eq:gru hthat}
\hat{\mathbf{h}_{t}} = \tanh \left( \mathbf{W} \left(\begin{array}{ccc} \mathbf{x}_{t} \\ r\odot \mathbf{h}_{t-1} \\ \end{array} \right) \right)
\end{equation} 
\begin{equation}\label{eq:gru ht}
\mathbf{h}_{t} = \mathbf{z}\odot \mathbf{h}_{t-1} + (1-\mathbf{z})\odot \hat{\mathbf{h}}_{t}
\end{equation}
where \begin{math} \hat{\mathbf{h}}_{t} \end{math} denotes new candidate state values. 

\subsubsection{Bidirectional Recurrent Neural Network}
A bidirectional RNN \cite{schuster1997bidirectional} performs a forward pass and a backward pass, which runs input data from \begin{math}t=T \end{math} to \begin{math} t=1 \end{math} and from \begin{math}t=1 \end{math} to \begin{math} t=T \end{math},  respectively. 
%This means, it has double amount of neurons as an one directional RNN.

For classification, the output of an RNN \begin{math} \mathbf{y}_{t} \end{math} can be passed to a fully-connected layer with softmax activation functions; this allows us to interpret the output as a probability.% This gives probabilities over each class label.

\subsubsection{Batch Normalization}
To train a deep neural network, the \textit{internal covariate shift} \cite{ioffe2015batch} slows down the training process. The  \textit{internal covariate shift} is the distribution of each layer's input changes during training, because the parameters in the previous layer are changing. To reduce the \textit{internal covariate shift}, we could whiten the layer activations, but this takes too much computation power. Batch normalization, a part of the neural network structure, approximates this process by standardizing the activations \begin{math} \mathbf{x} \end{math} using a statistical estimate of the mean \begin{math}  \widehat{\mathrm{E}}(\mathbf{x}) \end{math} and standard deviation \begin{math} \widehat{\mathrm{Var}}(\mathbf{x}) \end{math} for each training mini-batch. It can be shown that
\begin{equation}\label{eq:bn}
\mathrm{BN}(\mathbf{x};\gamma,\beta)=\gamma \frac{\mathbf{x}-\widehat{\mathrm{E}}(\mathbf{x})}{ \sqrt{\widehat{\mathrm{Var}}(\mathbf{x})+\epsilon}}+\beta
\end{equation}
where \begin{math}  \gamma \in \mathbb{R}^{d}  \end{math} and \begin{math}  \beta \in \mathbb{R}^{d} \end{math} are scale and shift parameters for the activation \begin{math}  \mathbf{x} \in \mathbb{R}^{n\times d}  \end{math}. With these, identity transformation for each activation could be presented.  \begin{math} \epsilon \in \mathbb{R}  \end{math} is a constant added as a regularization parameter for numerical stability. The division in  Eq.~(\ref{eq:bn}) is performed element-wise. \begin{math}  \gamma \end{math} and \begin{math}  \beta \end{math} are learned during training and fixed during inference. 

\subsubsection{Proposed RNN Structure}
\begin{figure}%[thpb]
	\centering
	\includegraphics[width=3.0 in]{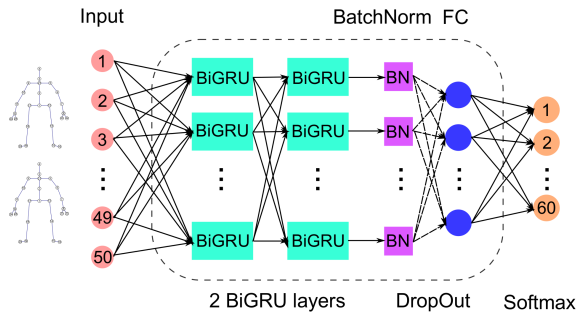}
	\caption{Proposed RNN structure using two bidirectional gated recurrent layers with batch normalization, dropout, one hidden fully-connected layer, and one output softmax layer. For clarity, the temporal recurrent structure of GRU cells is not shown here.}
	\label{fig:rnnStructure}
\end{figure}

For skeleton-based action recognition tasks, the data set consists of the 3D coordinates of a number of body joints. We feed this information, together with action labels, to an RNN. This RNN network has two bidirectional layers, each of which consists of 300 GRU cells.
After the recurrent layers follows the batch normalization layer, which standardizes the activations from the RNN layer. Then the normalized activations flow to the next fully-connected layer with 600 rectified linear unit (ReLU) \cite{nair2010rectified} activation functions. During training, in each iteration the network randomly drops out 25\% of the neurons between the batch normalization layer and the next fully-connected layer to reduce overfitting.
Lastly, a softmax layer maps the compressed motion information (features) to 60 action classes. Fig.~\ref{fig:rnnStructure} shows the structure of this RNN network.  

\par

To highlight the improvements of this final proposed model, we compare our approach to simpler models.
These models are a standard RNN; an LSTM-RNN; LSTM plus batch normalization (``LSTM-BN''), LSTM-BN with dropout (``LSTM-BN-DP''),  GRU-BN-DP, and a bidirectional GRU-BN-DP which we call ``BI-GRU-BN-DP''.
All these models have one recurrent layer.  The next complexity is adding an extra layer of hidden units to the last model (``2 layer BI-GRU-BN-DP'').  Finally, we add another fully-connected layer on top, before the softmax layer, and call this model ``2 layer BI-GRU-BN-DP-H''.
Sec.~\ref{sec:Experiment} discusses the results of all models.

\subsection{Convolutional Neural Network}
\label{sec:Convolutional Neural Network}
To process RGB videos, we choose to use the 3D-CNN model from\cite{tran2015learning}, as it shows promising performances on 2D video action recognition tasks. We believe that 3D convolution nets are more suitable for learning features from videos than 2D convolution nets.\par
\begin{figure}%[thpb]
	\centering
	\includegraphics[width=2.0 in]{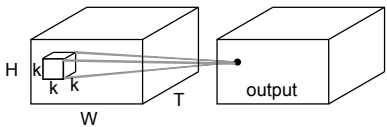}
	\caption{Illustration of 3D convolution operation on a video volume resulting in another volume. H and W are height and width of a frame, T means the maximal time step of a video. k denotes the size of a kernel. }
	\label{fig:cnnCube}
\end{figure}

 
2D convolution generates a series of 2D feature maps from images. Inspired by this, a 3D convolution processes frame clips, where the third dimension is time step, which results in a series of 3D feature volumes, as shown in Fig.~\ref{fig:cnnCube}. This compressed representation contains spatiotemporal information from the video clips. To learn a rich amount of features, multiple layers of convolution and max-pooling operations are stacked into one model. 

\par
\begin{figure}%[thpb]
	\centering
	\includegraphics[width=3.3 in]{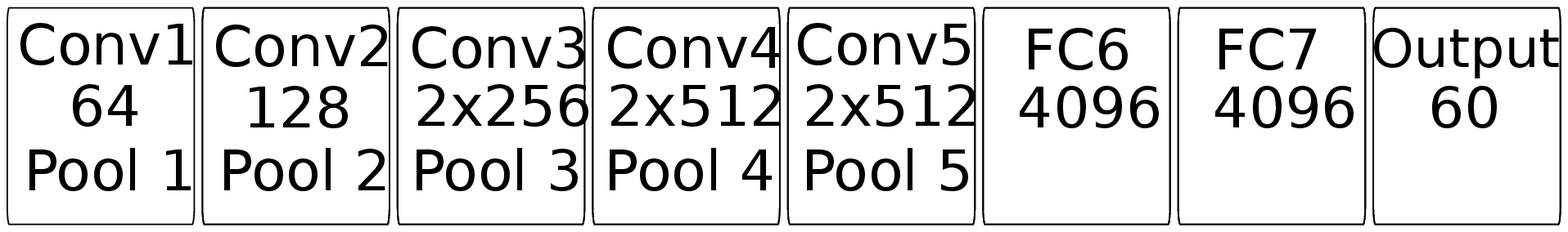}
	\caption{3D-CNN structure (C3D): 8 convolution, 5 max-pooling, 2 fully-connected layers, and 1 softmax output layer. All convolution kernels are of size $ 3 \times 3 \times 3 $ with stride 1. Number of filters are shown in each box. All pooling kernels are $ 2 \times 2 \times 2 $, except the first one, which is $ 1 \times 2 \times 2 $. Each fully-connected layer has 4096 units.}
	\label{fig:cnnStructure}
\end{figure}

 
To be specific, the 3D-CNN model \cite{tran2015learning}, which we choose, has five convolutional groups, each group has one or two convolutional layers and one max-pooling layer, two fully-connected layers, and one softmax output layer. The details of this model are presented in Fig.~\ref{fig:cnnStructure}. \par
 We finetune this model with pretrained parameters \cite{tran2015learning} on Sports-1M Dataset, which has approximately one million YouTube videos. This reduces overfitting and demands less training time on the current dataset.
\par

\subsection{Two-stream RNN/CNN}
\label{sec:Two-stream RNN/CNN}
As having the proposed RNN structure for the skeleton data and the 3D-CNN model for the RGB videos, we want to combine the strengths of RNN and CNN nets. To improve the performance, we propose two fusion models, decision fusion and feature fusion. \par
\subsubsection{Decision Fusion}
In the case of decision fusion, we use a simple but efficient voting method, inspired by majority voting. As a result of having only two classifiers, we cannot apply majority voting. Instead, the fusion method predicts based on voting confidence. \par
We first split the dataset into training, validation and testing. The same training set is used to train the RNN and CNN nets. The validation set is then used to find the best parameters, trust weights \begin{math} w_{r}\end{math} and \begin{math} w_{c}\end{math} for the voting method. We initialize the trust weights with equal values, which means \begin{math} w_r=1.00\end{math} and \begin{math} w_c=1.00\end{math} for both RNN and CNN classifiers.  Afterwards, we compare the confidences, which are the highest probabilities of softmax output from both classifiers for each prediction. The more confident one wins:
\begin{equation}\label{eq:decisionFusion}
    y(x_i)= 
\begin{cases}
    y_{r}(x_i),& \text{if } w_{r} \times y_{r}(x_i) > w_{c} \times y_{c}(x_i)\\
    y_{c}(x_i),              & \text{otherwise}
\end{cases}
\end{equation} 
where \begin{math}y(x_i) \end{math} is the fused prediction for sample \begin{math}x_i \end{math}; \begin{math}y_{r} \end{math} and \begin{math}y_{c} \end{math} denote RNN and CNN prediction, respectively.

Based on this concept, we develop a way to fuse the predictions from RNN and CNN. We evaluate the performance with the validation dataset and search for the best trust weights for decision fusion. 
Having only two parameters \begin{math} w_r\end{math} and \begin{math} w_c\end{math}, only little tuning is needed.

\subsubsection{Feature Fusion}    
Another way to combine these two neural networks is feature fusion. \par
We first train the RNN and CNN models on the training dataset. As in training, neural nets can learn discriminant information from raw data. Thus, we use the trained RNN model to extract temporal features from 3D skeleton data and use the trained CNN model to learn spatiotemporal features from RGB videos. Both features come from the first fully-connected layer in each model. The features are concatenated, L2 normalized, and eventually, fed to a linear SVM classifier.  \par
\begin{figure}%[thpb]
	\centering
	\includegraphics[width=3 in]{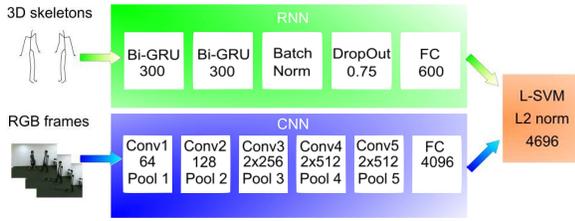}
	\caption{Two-stream RNN/CNN structure: The RNN stream is fed with the 3D coordinates of two human skeletons as input, then followed by two bidirectional gated recurrent layer with 300 units in each direction. The output from recurrent layers is later batch-normalized. 
	  Dropout is only enabled during training with 75\% keep probability. The RNN features come from the fully-connected layer. The CNN stream is fed with RGB clips (16 frames as a clip) and consists of five convolution groups and a fully-connected layer (fc-6). The CNN features are extracted from the fc-6 layer and later combined with RNN features, then L2-normalized, and fed to a linear SVM, which predicts the actions.}
	\label{fig:net_structure}
\end{figure}

 
The SVM parameter $C$  is found using the validation dataset. Then the model is tested on the test set. The feature fusion structure for two streams of RNN and CNN features is presented in Fig.~\ref{fig:net_structure}.

\section{Experiments}
\label{sec:Experiment}

The models introduced in the previous section are evaluated in the experiments. The dataset is first introduced in this section, then the setups and parameter settings for the experiments are illustrated.  We compare the results of the proposed models with the current best methods. In the end, we analyze and discuss the problems related to deep learning methods for 3D action recognition. \par
  
\subsection{NTU RGB+D Dataset \cite{Shahroudy_2016_CVPR}} 
\label{sec:NTU RGB+D Dataset}
The proposed approaches are evaluated on the NTU RGB+D dataset\cite{Shahroudy_2016_CVPR}, which we know as the current largest publicly available 3D action recognition dataset. 
The dataset consists of more than 56k action videos and 4 million frames, which were collected by 3 Kinect V2 cameras from 40 distinct subjects, and divided into 60 different action classes including 40 daily (drinking, eating, reading, etc.), 9 health-related (sneezing, staggering, falling down, etc.), and 11 mutual (punching, kicking, hugging, etc.)\ actions. It has four major data modalities provided by the Kinect sensor: 3D coordinates of 25 joints for each person (skeleton), RGB frames, depth maps, and IR sequences. In this paper, we use the first two modalities, since they are the two most informative modalities. 
 
The large intra-class and view point variations make this dataset challenging. However, the large amount of action samples makes it highly suitable for data-driven methods.\par

This dataset has two standard evaluation criteria \cite{Shahroudy_2016_CVPR}. The first one is a cross-subject test, in which half of the subjects are used for training and the other half are used for testing. The second one is a cross-view test, in which two viewpoints are used for training and one is excluded for evaluation. \par

\subsection{Implementation details} 
\label{sec:Implementation details}
In our experiments, the implementation consists of RNN, CNN, and Fusion. For all these models we use the same training, validation and testing splits. The validation set is composed of 10\% of the subjects in the training set in  \cite{Shahroudy_2016_CVPR}. The remaining subjects in the training set \cite{Shahroudy_2016_CVPR} make up the training set. 

\subsubsection{RNN Implementation}
In the RNN experiments, we have two human skeletons as input, each skeleton has 25 3D coordinates. Since the longest time step is 300, we pad all the action sequences to a length of 300. The dimension of each action sample is  300\,(time steps)\,\begin{math} \times \end{math}\,150\,(coordinates). \par 

We use TensorFlow\cite{abadi2016tensorflow} with TFlearn\cite{tflearn2016}  and run the experiments on either one NVIDIA GTX 1080 GPU or one NVIDIA GTX TITAN X GPU. We train the network using RMSprop \cite{tieleman2012lecture} optimizer and set learning rate as 0.001, decay as 0.9, and momentum as 0. We train the network from scratch using mini-batches of 1000 sequences for one-layer models and use mini-batches of 650 sequences for two-layer models. For all RNN nets, we use 300 neurons for each single-directional layer, double the amount of neurons for bidirectional layers,  and we use a 75\% keep probability for dropout. For batch normalization, we initialize \begin{math} \gamma \end{math} as 1.0, \begin{math} \beta \end{math} as 0.0, and set \begin{math} \epsilon \end{math} as \begin{math} 1\times 10^{-5} \end{math}. The estimated means and variances are fixed during inference.\par
As a comparison, the mentioned parameters are the same for all proposed RNN models, only the structure changes. \par

\subsubsection{CNN Implementation}
We use the 3D-CNN model\cite{tran2015learning} in Caffe\cite{jia2014caffe} and train it on RGB frames from the NTU RGB+D dataset, with pretrained parameters\cite{tran2015learning} from the Sport1M dataset. 
From RGB videos, we extract the frames, crop and resize them from $1920\times1080$\,pixels to $320\times240$\,pixels \cite{tran2015learning}. Videos are split into non-overlapped 16-frame clips.\par
We refer to the input of CNN model as a size of \begin{math} c\times t \times h \times w \end{math}, where \begin{math} c\end{math} is the number of channels, \begin{math} t\end{math} is the number of time steps, \begin{math}h \end{math} and \begin{math} w\end{math} are the height and width of the frame, respectively. The network takes video clips as input and predicts the 60 action labels which belong to the 60 different actions. It further resizes the input frames to 128\begin{math} \times \end{math}171\,pixel resolution. The input dimensions are 3\begin{math} \times \end{math}16\begin{math} \times \end{math}128\begin{math} \times \end{math}171\,pixel. During training we use jittering on the input clips by random cropping them into 3\begin{math} \times \end{math}16\begin{math} \times \end{math}112\begin{math} \times \end{math}112\,pixel.  We fine-tune the network with stochastic gradient descent optimizer using mini-batches of 44 clips, with initial learning rate of 0.0001. The learning rate is then reduced by half, when no training progress was observed \cite{tieleman2012lecture}. The training stopped after around 20 epochs.

For video-based prediction, the model averages the predictions over all 16-frame clips split from the same video and provides the final prediction for the input video. A similar idea is applied for extracting features from fc-6 layer, which averages the 4096-dimensional feature vectors over all clips in the same video, resulting in one 4096-dimensional vector for each video.\par

\subsubsection{Fusion Implementation}
We fuse the best RNN structure, 2 layer BI-GRU-BN-DP-H, with the 3D-CNN model, first using decision fusion, then using feature fusion.

%The implementation for decision fusion, voting based on confidence, is implemented in python. 
For decision fusion, we first extract the softmax output, then search for the fusion parameters, trust weight \begin{math} w_r\end{math} and \begin{math} w_c\end{math} for RNN and CNN from the validation split. The parameters are \begin{math} w_r=1.00\end{math} and \begin{math} w_c=2.88\end{math} for the cross subject setup, and \begin{math} w_r=1.00\end{math} and \begin{math} w_c=3.02\end{math} for the cross view setup.\par  

For feature fusion, we extract the RNN features (600\,dimensions) from the fully-connected layer, and extract CNN features (4096\,dimensions) from the fc-6 layer \cite{tran2015learning}. We then concatenate them into one feature array (4,696\,dimensions) and apply L2 normalization. In the end, we have normalized RNN/CNN features from training, validation, and testing splits. We use training and validation splits to find the optimal value of \begin{math} C\end{math} for linear SVM\cite{scikit-learn} model. For both cross-subject and cross-view setups, we find that \begin{math}C=8.0\end{math} gives the best validation accuracy. \par 
Among all the models in this paper, feature fusion model shows the best testing results. We refer to this model as a two-stream RNN/CNN structure as shown in Fig.~\ref{fig:net_structure}.\par

\subsection{Experimental Results and Analysis}
\label{sec:Experimental Results and Analysis}
The evaluation results are shown in Tab.~\ref{table:results}. The first 16 rows are skeleton-based methods. The 3D-CNN model (17th row) uses RGB videos as input. The decision fusion (18th row) and feature fusion (19th row) models use the best RNN structure, which is the 2 Layer BI-GRU-BN-DP-H (16th row), and the 3D-CNN (17th row) model.  

\begin{figure}%[thpb]
	\centering
	\includegraphics[width=3.3 in]{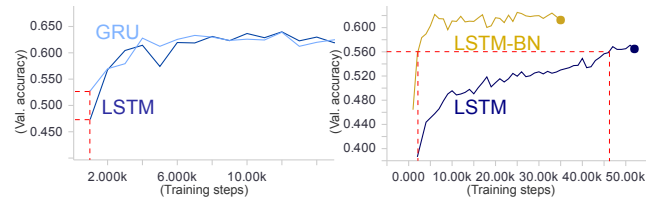}
	\caption{The left figure shows: For 1k training steps, the model with GRU cells takes 20\,m 31\,s, reaches 52.75\% validation accuracy, while the same model with LSTM cells takes 24\,m 27\,s and only has 47.33\% validation accuracy. The right figure shows: For validation accuracy 56\%, LSTM-BN takes 2\,k training steps, while LSTM needs 46\,k training steps. Batch normalization makes the model converge 13 times faster.}
	\label{fig:gru_bn_faster}
\end{figure}

Tab.~\ref{table:results} shows that our RNN structure, the 1 Layer LSTM-BN, already outperforms the baseline method part-aware LSTM reported in \cite{Shahroudy_2016_CVPR} because batch normalization improves the LSTM model. %. And 1 Layer LSTM-BN model needs 13 times fewer training steps than a normal 1 Layer LSTM model, as shown in Fig.~\ref{fig:gru_bn_faster} right. The batch normalization layer also improves the prediction accuracy on both setups.\par  
Adding a dropout procedure reduces overfitting and further improves the results (rows 11, 12). %ropout \cite{srivastava2014dropout} can further keep the network from 
From rows 12 and 13 we can see that the performances of LSTM and GRU cells are similar\cite{chung2014empirical}. GRU is better in the cross-subject test and LSTM is better in the cross-view test. On the other hand, GRU is faster than LSTM both in computational speed and converge rate. As presented in Fig.~\ref{fig:gru_bn_faster} left, for 1k training steps, the same model performs 5.42\% more accurately and takes 20\% less computational time when using GRU cells than when using LSTM cells. 

\begin{table}%[thpb]
\centering
\begin{tabular}{ p{0.2cm}  p{3.9cm}  p{1.5cm}  p{1.2cm} }\hlinewd{1pt}
Nr. & Method 														&	cross subject		& 	cross view \\ \hline
01 &Skeleton Quads\cite{evangelidis2014skeletal,Shahroudy_2016_CVPR} & 38.62\% & 41.36\% \\ 
02 &Lie Group\cite{vemulapalli2014human,Shahroudy_2016_CVPR} & 50.08\% & 52.76\% \\ 
03 &FTP Dynamic Skeletons\cite{hu2015jointly,Shahroudy_2016_CVPR} & 60.23\% & 65.22\% \\ 
04 &HBRNN-L\cite{du2015hierarchical,Shahroudy_2016_CVPR} & 59.07\% & 63.97\% \\ 
05 &Deep RNN\cite{Shahroudy_2016_CVPR} & 56.29\% & 64.09\% \\
06 &Deep LSTM\cite{Shahroudy_2016_CVPR} & 60.69\% & 67.29\% \\ 
07 &Part-aware LSTM\cite{Shahroudy_2016_CVPR} & 62.93\% & 70.27\% \\ 
08 &ST-LSTM (Tree) + Trust Gate\cite{liu2016spatio}&\textcolor{BlueViolet}{\textbf{69.2\%}}&\textcolor{BlueViolet}{\textbf{77.7\%}}\\ \hline
09 &1 Layer RNN												&	18.74\% &	20.27\%	 \\ 
10 &1 Layer LSTM												&	60.99\% &	64.68\%		\\ 
11 &1 Layer LSTM-BN											&	64.07\% &	71.86\%		\\ 
12 &1 Layer LSTM-BN-DP									&	64.69\% &	73.48\%		\\ 
13 &1 Layer GRU-BN-DP									&	65.21\% &	70.36\%		\\ 
14 &1 Layer BI-GRU-BN-DP								&	64.78\% &	73.12\%		\\ 
15 &2 Layer BI-GRU-BN-DP				&	66.21\% &72.46\%\\
16 &\textcolor{black}{2 Layer BI-GRU-BN-DP-H}				&	\textcolor{violet}{\textbf{70.70\%}} &\textcolor{violet}{\textbf{80.23\%}}\\ \hline
17 &\textcolor{black}{3D-CNN}\cite{tran2015learning}	&	\textcolor{violet}{\textbf{79.75\%}} &	\textcolor{violet}{\textbf{83.95\%}}\\ 
18 &Decision Fusion								&	82.05\% &	86.68\%		\\ 
19 &Feature Fusion								&	\textcolor{black}{\textbf{83.74\%}} &	\textcolor{black}{\textbf{93.65\%}}		\\ \hlinewd{1pt}
\end{tabular}
\caption{Comparison of testing accuracies on the NTU RGB+D dataset. The first 8 rows are the results reported from other papers. Rows 9 to 19 are results of the methods evaluated in this paper. The first 16 rows are skeleton-based methods. Our 2 Layer BI-GRU-BN-DP-H model outperforms other methods on both evaluation protocols. In addition, the feature fusion model further boosts accuracy by more than 13\% on both settings.}
\label{table:results}
\end{table}

The addition of the extra fully-connected layer brings another significant improvement. This increases the complexity of the neural network, which helps the model capture more inherent features from the 3D skeleton data \cite{cybenko1989approximation}.
The recurrent layers before the fully-connected layer can be seen as a temporal feature extractor, which compact input information (dimension \begin{math} 300 \times 150 \end{math}) into 600 dimensions.
The latter part of the RNN structure can be considered as a classifier learning to map these 600-dimensional features to 60 different action categories.
%Both these too parts are quite important to our proposed RNN structure.\par 
Altogether, our novel RNN model, 2 Layer BI-GRU-BN-DP-H, outperforms all the other skeleton-based models including ST-LSTM (Tree traversal) + Trust Gate \cite{liu2016spatio}.% and reaches state-of-the-art performance.\par

Then, we use the RGB video data to train the 3D-CNN model.
%
%Improvements are not ended here, we newly add  RGB videos from the NTU dataset to the experiments. We first feed the RGB frames to the 3D-CNN model, train the model, and compare the results with our best RNN model. 
We use the voting method based on confidence to fuse the 2 Layer BI-GRU-BN-DP-H and 3D-CNN model. 
In the next step, we utilize a linear SVM \cite{tran2015learning} to fuse the fc-6 features from the CNN and the fc features from the RNN. This further improves results by over 13\% in comparison to our best RNN, and by more than 14\% compared to literature models \cite{Shahroudy_2016_CVPR,liu2016spatio}. This boosting is due to the features from RNN and CNN model being highly complementary. The RNN model uses 50 3D coordinates for two human bodies over 300 time steps, and learns to find the long-term motion pattern. Whereas the CNN model has 2D RGB frames, which additionally has spatiotemporal information about objects, such as cups, pens, and books. However, the CNN model can only memorize information for 16 time steps long---longer memorization is prohibited by GPU memory limitations. These facts make the features from RNN and CNN model highly complementary, as the testing results show in row 16, 17, and 19 in Tab.~\ref{table:results}.

\subsection{Discussion}
\label{sec:Discussion}

\begin{figure}%[thpb]
	\centering
	\includegraphics[width=3.4 in]{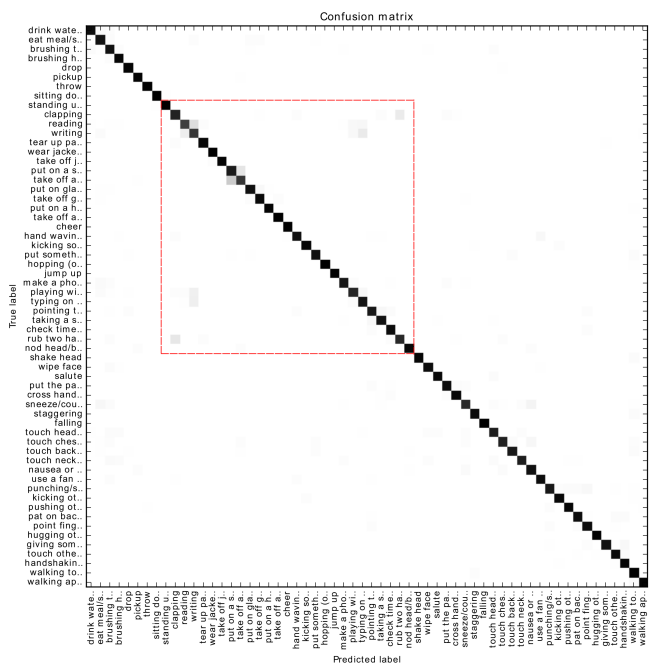}
	\caption{The confusion matrix of the results from the feature fusion method using the cross-view test. The rectangle area is shown in more detail in Fig.~\ref{fig:cm_clappingRub_01}.}
	\label{fig:cm_wb_01}
\end{figure}

\begin{figure}%[thpb]
	\centering
	\includegraphics[width=3.45 in]{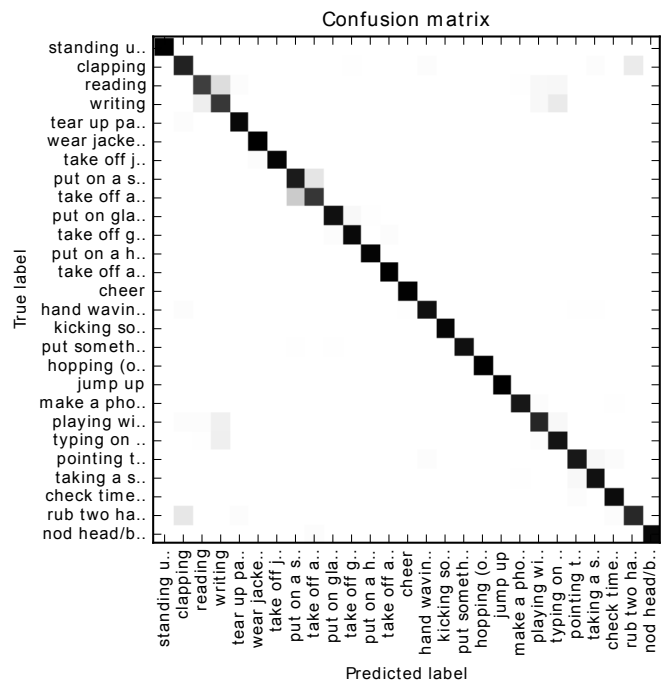}
	\caption{A part of the confusion matrix shown in Fig.~\ref{fig:cm_wb_01}. As the figure shows, action pairs such as reading vs.\ writing, putting on a shoe vs.\ taking off a shoe, and rubbing two hands vs.\ clapping, are relatively confusing to the two-stream RNN/CNN structure.}
	\label{fig:cm_clappingRub_01}
\end{figure}

\begin{figure}
	\centering
	\includegraphics[width=3.4in]{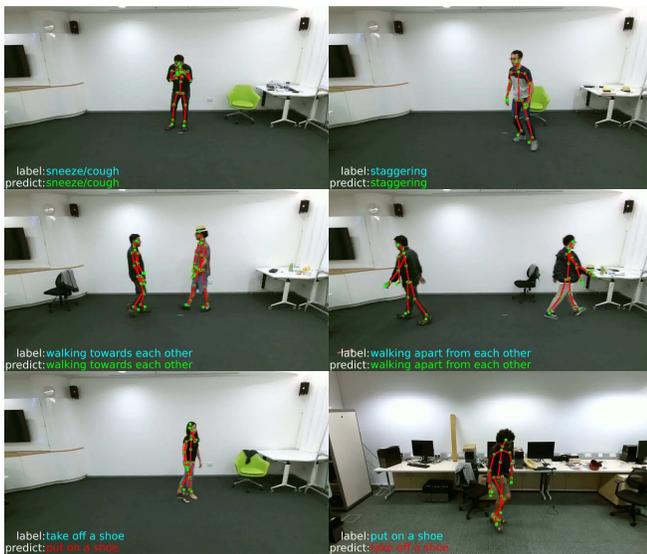}
	\caption[Some classified action samples]{Some correctly classified action samples (first four frames with green predictions) and some mis-classified action samples (last two frames with red predictions). These samples are randomly picked from the feature fusion model cross view testing results.}
	\label{fig:frame3}
\end{figure}

To better analyze and improve the performance of the model, we take a closer look at actions that are highly confusing to the two-stream RNN/CNN structure. As presented in Fig.~\ref{fig:cm_wb_01} and \ref{fig:cm_clappingRub_01}, such action pairs include reading vs.\ writing, putting on a shoe vs.\ taking off a shoe, and rubbing two hands vs.\ clapping. These actions are shown in a video at \url{https://www.youtube.com/watch?v=G0PXKCEgIoA}. Fig.~\ref{fig:frame3} shows some classified action samples.

There could be several reasons for this observation. First, these actions are sometimes inherent confusing. Secondly, there are flaws in the data. Kinect depth information, from which the NTU skeleton data is created, is quite noisy \cite{shotton2013real,mallick2014characterizations}.  Correspondingly, the 3D skeleton data used in our RNNs are also quite noisy \cite{liu2016spatio}. RGB videos data are more accurate and stable, but single frames carry no 3D information. Thirdly, the 3D-CNN model\cite{tran2015learning} is trained with small video clips, which are 16 time steps long. The CNN model is adapted to find only short-term temporal features in these clips. As GPU memory and computing power increase, the model could also be adapted to find long-term temporal features in each whole video. Lastly, although, the RNN model can memorize the whole action sequences and give final predictions, it has no information about the appearances and movements of surrounding objects, which could be discriminative information for the classification task.        
%\todo[inline]{Patrick says: I don't understand what you want to say in the next sentence.}Besides, the 3D-CNN \cite{tran2015learning} model, even the LRCN (CNN + LSTM) \cite{donahue2015long} model, is only trained with 16-frame long clips. It could be more powerful, if these models could handle much longer input sequence such as 300 time stheps in the skeleton-based RNN structures. \par   

\section{Conclusion and Future Work}
In this paper, we propose a novel RNN structure for 3D skeletons that achieves state-of-the-art performance on the largest 3D action recognition dataset. The proposed RNN model can also be trained 13 times faster and saves 20\% computational power on each training step. Additionally, the RGB videos from the same dataset are used to finetune a 3D-CNN model. In the end, an efficient fusion structure, two-stream RNN/CNN, is introduced to fuse the capabilities of both RNN and CNN models. The results of this method are 13\% higher than using the proposed RNN alone, and 14\% higher than the best published result in the literature. In the future, we want to consider using the other sensor modalities such as depth maps and IR sequences and see what is the best architecture to fuse all these modalities. \par 

%\section{Acknowledgements}
%This research is supported in part by the Robotics and Mechatronics Center (RMC) Institute of German Aerospace Center (DLR). The research has been carried out at RMC and TUM. We would like to thank Nanyang Technological University for providing the large scale dataset and thank all the teams involved for developing these awesome deep learning frameworks.  We are particularly grateful to Grady Jensen for his assistance.

%\addtolength{\textheight}{-1.9cm}   % This command serves to balance the column lengths

\bibliography{IEEEabrv,mybib}

\begin{thebibliography}{10}
\providecommand{\url}[1]{#1}
\csname url@samestyle\endcsname
\providecommand{\newblock}{\relax}
\providecommand{\bibinfo}[2]{#2}
\providecommand{\BIBentrySTDinterwordspacing}{\spaceskip=0pt\relax}
\providecommand{\BIBentryALTinterwordstretchfactor}{4}
\providecommand{\BIBentryALTinterwordspacing}{\spaceskip=\fontdimen2\font plus
\BIBentryALTinterwordstretchfactor\fontdimen3\font minus
  \fontdimen4\font\relax}
\providecommand{\BIBforeignlanguage}[2]{{%
\expandafter\ifx\csname l@#1\endcsname\relax
\typeout{** WARNING: IEEEtran.bst: No hyphenation pattern has been}%
\typeout{** loaded for the language `#1'. Using the pattern for}%
\typeout{** the default language instead.}%
\else
\language=\csname l@#1\endcsname
\fi
#2}}
\providecommand{\BIBdecl}{\relax}
\BIBdecl

\bibitem{li2010action}
W.~Li, Z.~Zhang, and Z.~Liu, ``Action recognition based on a bag of {3D}
  points,'' in \emph{2010 IEEE Computer Society Conference on Computer Vision
  and Pattern Recognition-Workshops}.\hskip 1em plus 0.5em minus 0.4em\relax
  IEEE, 2010, pp. 9--14.

\bibitem{Shahroudy_2016_CVPR}
A.~Shahroudy, J.~Liu, T.-T. Ng, and G.~Wang, ``{NTU RGB+D}: A large scale
  dataset for {3D} human activity analysis,'' in \emph{Proceedings of the IEEE
  Conference on Computer Vision and Pattern Recognition}, 2016.

\bibitem{du2015hierarchical}
Y.~Du, W.~Wang, and L.~Wang, ``Hierarchical recurrent neural network for
  skeleton based action recognition,'' in \emph{Proceedings of the IEEE
  Conference on Computer Vision and Pattern Recognition}, 2015, pp. 1110--1118.

\bibitem{liu2016spatio}
J.~Liu, A.~Shahroudy, D.~Xu, and G.~Wang, ``Spatio-temporal {LSTM} with trust
  gates for 3d human action recognition,'' \emph{arXiv preprint
  arXiv:1607.07043}, 2016.

\bibitem{zhu2016co}
W.~Zhu, C.~Lan, J.~Xing, W.~Zeng, Y.~Li, L.~Shen, and X.~Xie, ``Co-occurrence
  feature learning for skeleton based action recognition using regularized deep
  lstm networks,'' \emph{arXiv preprint arXiv:1603.07772}, 2016.

\bibitem{wang20143d}
K.~Wang, X.~Wang, L.~Lin, M.~Wang, and W.~Zuo, ``3d human activity recognition
  with reconfigurable convolutional neural networks,'' in \emph{Proceedings of
  the 22nd ACM international conference on Multimedia}.\hskip 1em plus 0.5em
  minus 0.4em\relax ACM, 2014, pp. 97--106.

\bibitem{wang2015action}
P.~Wang, W.~Li, Z.~Gao, J.~Zhang, C.~Tang, and P.~O. Ogunbona, ``Action
  recognition from depth maps using deep convolutional neural networks,'' 2015.

\bibitem{tran2015learning}
D.~Tran, L.~Bourdev, R.~Fergus, L.~Torresani, and M.~Paluri, ``Learning
  spatiotemporal features with {3D} convolutional networks,'' in \emph{2015
  IEEE International Conference on Computer Vision (ICCV)}.\hskip 1em plus
  0.5em minus 0.4em\relax IEEE, 2015, pp. 4489--4497.

\bibitem{evangelidis2014skeletal}
G.~Evangelidis, G.~Singh, and R.~Horaud, ``Skeletal quads: Human action
  recognition using joint quadruples,'' in \emph{International Conference on
  Pattern Recognition}, 2014, pp. 4513--4518.

\bibitem{vemulapalli2014human}
R.~Vemulapalli, F.~Arrate, and R.~Chellappa, ``Human action recognition by
  representing {3D} skeletons as points in a lie group,'' in \emph{Proceedings
  of the IEEE Conference on Computer Vision and Pattern Recognition}, 2014, pp.
  588--595.

\bibitem{hu2015jointly}
J.-F. Hu, W.-S. Zheng, J.~Lai, and J.~Zhang, ``Jointly learning heterogeneous
  features for {RGB-D} activity recognition,'' in \emph{Proceedings of the IEEE
  conference on computer vision and pattern recognition}, 2015, pp. 5344--5352.

\bibitem{rahmani2016histogram}
H.~Rahmani, A.~Mahmood, D.~Huynh, and A.~Mian, ``Histogram of oriented
  principal components for cross-view action recognition,'' 2016.

\bibitem{gaglio2015human}
S.~Gaglio, G.~L. Re, and M.~Morana, ``Human activity recognition process using
  3-d posture data,'' \emph{IEEE Transactions on Human-Machine Systems},
  vol.~45, no.~5, pp. 586--597, 2015.

\bibitem{chen2013real}
C.~Chen, K.~Liu, and N.~Kehtarnavaz, ``Real-time human action recognition based
  on depth motion maps,'' \emph{Journal of real-time image processing}, pp.
  1--9, 2013.

\bibitem{ni2013rgbd}
B.~Ni, G.~Wang, and P.~Moulin, ``Rgbd-hudaact: A color-depth video database for
  human daily activity recognition,'' in \emph{Consumer Depth Cameras for
  Computer Vision}.\hskip 1em plus 0.5em minus 0.4em\relax Springer, 2013, pp.
  193--208.

\bibitem{sung2012unstructured}
J.~Sung, C.~Ponce, B.~Selman, and A.~Saxena, ``Unstructured human activity
  detection from rgbd images,'' in \emph{Robotics and Automation (ICRA), 2012
  IEEE International Conference on}.\hskip 1em plus 0.5em minus 0.4em\relax
  IEEE, 2012, pp. 842--849.

\bibitem{wang2012mining}
J.~Wang, Z.~Liu, Y.~Wu, and J.~Yuan, ``Mining actionlet ensemble for action
  recognition with depth cameras,'' in \emph{Computer Vision and Pattern
  Recognition (CVPR), 2012 IEEE Conference on}.\hskip 1em plus 0.5em minus
  0.4em\relax IEEE, 2012, pp. 1290--1297.

\bibitem{xia2012view}
L.~Xia, C.-C. Chen, and J.~Aggarwal, ``View invariant human action recognition
  using histograms of 3d joints,'' in \emph{2012 IEEE Computer Society
  Conference on Computer Vision and Pattern Recognition Workshops}.\hskip 1em
  plus 0.5em minus 0.4em\relax IEEE, 2012, pp. 20--27.

\bibitem{bloom2013dynamic}
V.~Bloom, V.~Argyriou, and D.~Makris, ``Dynamic feature selection for online
  action recognition,'' in \emph{International Workshop on Human Behavior
  Understanding}.\hskip 1em plus 0.5em minus 0.4em\relax Springer, 2013, pp.
  64--76.

\bibitem{lin2012human}
Y.-C. Lin, M.-C. Hu, W.-H. Cheng, Y.-H. Hsieh, and H.-M. Chen, ``Human action
  recognition and retrieval using sole depth information,'' in
  \emph{Proceedings of the 20th ACM international conference on
  Multimedia}.\hskip 1em plus 0.5em minus 0.4em\relax ACM, 2012, pp.
  1053--1056.

\bibitem{zhang2012privacy}
C.~Zhang, Y.~Tian, and E.~Capezuti, ``Privacy preserving automatic fall
  detection for elderly using rgbd cameras,'' in \emph{International Conference
  on Computers for Handicapped Persons}.\hskip 1em plus 0.5em minus 0.4em\relax
  Springer, 2012, pp. 625--633.

\bibitem{oreifej2013hon4d}
O.~Oreifej and Z.~Liu, ``Hon4d: Histogram of oriented 4d normals for activity
  recognition from depth sequences,'' in \emph{Proceedings of the IEEE
  Conference on Computer Vision and Pattern Recognition}, 2013, pp. 716--723.

\bibitem{koppula2013learning}
H.~S. Koppula, R.~Gupta, and A.~Saxena, ``Learning human activities and object
  affordances from rgb-d videos,'' \emph{The International Journal of Robotics
  Research}, vol.~32, no.~8, pp. 951--970, 2013.

\bibitem{negin2013decision}
F.~Negin, F.~{\"O}zdemir, C.~B. Akg{\"u}l, K.~A. Y{\"u}ksel, and
  A.~Er{\c{c}}il, ``A decision forest based feature selection framework for
  action recognition from rgb-depth cameras,'' in \emph{International
  Conference Image Analysis and Recognition}.\hskip 1em plus 0.5em minus
  0.4em\relax Springer, 2013, pp. 648--657.

\bibitem{wei2013concurrent}
P.~Wei, N.~Zheng, Y.~Zhao, and S.-C. Zhu, ``Concurrent action detection with
  structural prediction,'' in \emph{Proceedings of the IEEE International
  Conference on Computer Vision}, 2013, pp. 3136--3143.

\bibitem{munaro2013evaluation}
M.~Munaro, S.~Michieletto, and E.~Menegatti, ``An evaluation of 3d motion flow
  and 3d pose estimation for human action recognition,'' in \emph{RSS
  Workshops: RGB-D: Advanced Reasoning with Depth Cameras}, 2013.

\bibitem{ellis2013exploring}
C.~Ellis, S.~Z. Masood, M.~F. Tappen, J.~J. Laviola~Jr, and R.~Sukthankar,
  ``Exploring the trade-off between accuracy and observational latency in
  action recognition,'' \emph{International Journal of Computer Vision}, vol.
  101, no.~3, pp. 420--436, 2013.

\bibitem{mansur2013inverse}
A.~Mansur, Y.~Makihara, and Y.~Yagi, ``Inverse dynamics for action
  recognition,'' \emph{IEEE transactions on cybernetics}, vol.~43, no.~4, pp.
  1226--1236, 2013.

\bibitem{yang2013rgb}
Z.~Yang, L.~Zicheng, and C.~Hong, ``Rgb-depth feature for 3d human activity
  recognition,'' \emph{China Communications}, vol.~10, no.~7, pp. 93--103,
  2013.

\bibitem{carletti2013recognition}
V.~Carletti, P.~Foggia, G.~Percannella, A.~Saggese, and M.~Vento, ``Recognition
  of human actions from rgb-d videos using a reject option,'' in
  \emph{International Conference on Image Analysis and Processing}.\hskip 1em
  plus 0.5em minus 0.4em\relax Springer, 2013, pp. 436--445.

\bibitem{kastaniotis2013gait}
D.~Kastaniotis, I.~Theodorakopoulos, G.~Economou, and S.~Fotopoulos,
  ``Gait-based gender recognition using pose information for real time
  applications,'' in \emph{Digital Signal Processing (DSP), 2013 18th
  International Conference on}.\hskip 1em plus 0.5em minus 0.4em\relax IEEE,
  2013, pp. 1--6.

\bibitem{liu2015coupled}
A.-A. Liu, W.-Z. Nie, Y.-T. Su, L.~Ma, T.~Hao, and Z.-X. Yang, ``Coupled hidden
  conditional random fields for rgb-d human action recognition,'' \emph{Signal
  Processing}, vol. 112, pp. 74--82, 2015.

\bibitem{huang2014sequential}
D.~Huang, S.~Yao, Y.~Wang, and F.~De~La~Torre, ``Sequential max-margin event
  detectors,'' in \emph{European conference on computer vision}.\hskip 1em plus
  0.5em minus 0.4em\relax Springer, 2014, pp. 410--424.

\bibitem{lillo2014discriminative}
I.~Lillo, A.~Soto, and J.~Carlos~Niebles, ``Discriminative hierarchical
  modeling of spatio-temporally composable human activities,'' in
  \emph{Proceedings of the IEEE Conference on Computer Vision and Pattern
  Recognition}, 2014, pp. 812--819.

\bibitem{yu2014discriminative}
G.~Yu, Z.~Liu, and J.~Yuan, ``Discriminative orderlet mining for real-time
  recognition of human-object interaction,'' in \emph{Asian Conference on
  Computer Vision}.\hskip 1em plus 0.5em minus 0.4em\relax Springer, 2014, pp.
  50--65.

\bibitem{wu2015watch}
C.~Wu, J.~Zhang, S.~Savarese, and A.~Saxena, ``Watch-n-patch: Unsupervised
  understanding of actions and relations,'' in \emph{Proceedings of the IEEE
  Conference on Computer Vision and Pattern Recognition}, 2015, pp. 4362--4370.

\bibitem{hu2016exemplar}
J.-F. Hu, W.-S. Zheng, J.~Lai, S.~Gong, and T.~Xiang, ``Exemplar-based
  recognition of human--object interactions,'' \emph{IEEE Transactions on
  Circuits and Systems for Video Technology}, vol.~26, no.~4, pp. 647--660,
  2016.

\bibitem{chen2015utd}
C.~Chen, R.~Jafari, and N.~Kehtarnavaz, ``Utd-mhad: A multimodal dataset for
  human action recognition utilizing a depth camera and a wearable inertial
  sensor,'' in \emph{Image Processing (ICIP), 2015 IEEE International
  Conference on}.\hskip 1em plus 0.5em minus 0.4em\relax IEEE, 2015, pp.
  168--172.

\bibitem{cheng2012human}
Z.~Cheng, L.~Qin, Y.~Ye, Q.~Huang, and Q.~Tian, ``Human daily action analysis
  with multi-view and color-depth data,'' in \emph{European Conference on
  Computer Vision}.\hskip 1em plus 0.5em minus 0.4em\relax Springer, 2012, pp.
  52--61.

\bibitem{zhang2012viewpoint}
Z.~Zhang, W.~Liu, V.~Metsis, and V.~Athitsos, ``A viewpoint-independent
  statistical method for fall detection,'' in \emph{Pattern Recognition (ICPR),
  2012 21st International Conference on}.\hskip 1em plus 0.5em minus
  0.4em\relax IEEE, 2012, pp. 3626--3630.

\bibitem{ofli2013berkeley}
F.~Ofli, R.~Chaudhry, G.~Kurillo, R.~Vidal, and R.~Bajcsy, ``Berkeley mhad: A
  comprehensive multimodal human action database,'' in \emph{Applications of
  Computer Vision (WACV), 2013 IEEE Workshop on}.\hskip 1em plus 0.5em minus
  0.4em\relax IEEE, 2013, pp. 53--60.

\bibitem{amiri2013non}
S.~M. Amiri, M.~T. Pourazad, P.~Nasiopoulos, and V.~C. Leung, ``Non-intrusive
  human activity monitoring in a smart home environment,'' in \emph{e-Health
  Networking, Applications \& Services (Healthcom), 2013 IEEE 15th
  International Conference on}.\hskip 1em plus 0.5em minus 0.4em\relax IEEE,
  2013, pp. 606--610.

\bibitem{wei2013modeling}
P.~Wei, Y.~Zhao, N.~Zheng, and S.-C. Zhu, ``Modeling 4d human-object
  interactions for event and object recognition,'' in \emph{2013 IEEE
  International Conference on Computer Vision}.\hskip 1em plus 0.5em minus
  0.4em\relax IEEE, 2013, pp. 3272--3279.

\bibitem{wang2014cross}
J.~Wang, X.~Nie, Y.~Xia, Y.~Wu, and S.-C. Zhu, ``Cross-view action modeling,
  learning and recognition,'' in \emph{Proceedings of the IEEE Conference on
  Computer Vision and Pattern Recognition}, 2014, pp. 2649--2656.

\bibitem{rahmani2014hopc}
H.~Rahmani, A.~Mahmood, D.~Q. Huynh, and A.~Mian, ``Hopc: Histogram of oriented
  principal components of 3d pointclouds for action recognition,'' in
  \emph{European Conference on Computer Vision}.\hskip 1em plus 0.5em minus
  0.4em\relax Springer, 2014, pp. 742--757.

\bibitem{liu2015multipe}
A.-A. Liu, Y.-T. Su, P.-P. Jia, Z.~Gao, T.~Hao, and Z.-X. Yang,
  ``Multipe/single-view human action recognition via part-induced multitask
  structural learning,'' \emph{IEEE transactions on cybernetics}, vol.~45,
  no.~6, pp. 1194--1208, 2015.

\bibitem{song2014body}
Y.~Song, J.~Tang, F.~Liu, and S.~Yan, ``Body surface context: A new robust
  feature for action recognition from depth videos,'' \emph{IEEE Transactions
  on Circuits and Systems for Video Technology}, vol.~24, no.~6, pp. 952--964,
  2014.

\bibitem{yun2012two}
K.~Yun, J.~Honorio, D.~Chattopadhyay, T.~L. Berg, and D.~Samaras, ``Two-person
  interaction detection using body-pose features and multiple instance
  learning,'' in \emph{2012 IEEE Computer Society Conference on Computer Vision
  and Pattern Recognition Workshops}.\hskip 1em plus 0.5em minus 0.4em\relax
  IEEE, 2012, pp. 28--35.

\bibitem{hu2013efficient}
T.~Hu, X.~Zhu, W.~Guo, and K.~Su, ``Efficient interaction recognition through
  positive action representation,'' \emph{Mathematical Problems in
  Engineering}, vol. 2013, 2013.

\bibitem{wolf2014evaluation}
C.~Wolf, E.~Lombardi, J.~Mille, O.~Celiktutan, M.~Jiu, E.~Dogan, G.~Eren,
  M.~Baccouche, E.~Dellandr{\'e}a, C.-E. Bichot \emph{et~al.}, ``Evaluation of
  video activity localizations integrating quality and quantity measurements,''
  \emph{Computer Vision and Image Understanding}, vol. 127, pp. 14--30, 2014.

\bibitem{bloom2014g3di}
V.~Bloom, V.~Argyriou, and D.~Makris, ``G3di: A gaming interaction dataset with
  a real time detection and evaluation framework,'' in \emph{Workshop at the
  European Conference on Computer Vision}.\hskip 1em plus 0.5em minus
  0.4em\relax Springer, 2014, pp. 698--712.

\bibitem{van2014dyadic}
C.~Van~Gemeren, R.~T. Tan, R.~Poppe, and R.~C. Veltkamp, ``Dyadic interaction
  detection from pose and flow,'' in \emph{International Workshop on Human
  Behavior Understanding}.\hskip 1em plus 0.5em minus 0.4em\relax Springer,
  2014, pp. 101--115.

\bibitem{lin2015deep}
L.~Lin, K.~Wang, W.~Zuo, M.~Wang, J.~Luo, and L.~Zhang, ``A deep structured
  model with radius--margin bound for 3d human activity recognition,''
  \emph{International Journal of Computer Vision}, pp. 1--18, 2015.

\bibitem{wang2015convnets}
P.~Wang, W.~Li, Z.~Gao, C.~Tang, J.~Zhang, and P.~Ogunbona, ``Convnets-based
  action recognition from depth maps through virtual cameras and
  pseudocoloring,'' in \emph{Proceedings of the 23rd ACM international
  conference on Multimedia}.\hskip 1em plus 0.5em minus 0.4em\relax ACM, 2015,
  pp. 1119--1122.

\bibitem{hochreiter1997long}
S.~Hochreiter and J.~Schmidhuber, ``Long short-term memory,'' \emph{Neural
  computation}, vol.~9, no.~8, pp. 1735--1780, 1997.

\bibitem{ioffe2015batch}
S.~Ioffe and C.~Szegedy, ``Batch normalization: Accelerating deep network
  training by reducing internal covariate shift,'' \emph{arXiv preprint
  arXiv:1502.03167}, 2015.

\bibitem{cho2014learning}
K.~Cho, B.~Van~Merri{\"e}nboer, C.~Gulcehre, D.~Bahdanau, F.~Bougares,
  H.~Schwenk, and Y.~Bengio, ``Learning phrase representations using {RNN}
  encoder-decoder for statistical machine translation,'' \emph{arXiv preprint
  arXiv:1406.1078}, 2014.

\bibitem{donahue2015long}
J.~Donahue, L.~Anne~Hendricks, S.~Guadarrama, M.~Rohrbach, S.~Venugopalan,
  K.~Saenko, and T.~Darrell, ``Long-term recurrent convolutional networks for
  visual recognition and description,'' in \emph{Proceedings of the IEEE
  Conference on Computer Vision and Pattern Recognition}, 2015, pp. 2625--2634.

\bibitem{simonyan2014two}
K.~Simonyan and A.~Zisserman, ``Two-stream convolutional networks for action
  recognition in videos,'' in \emph{Advances in Neural Information Processing
  Systems}, 2014, pp. 568--576.

\bibitem{ji20133d}
S.~Ji, W.~Xu, M.~Yang, and K.~Yu, ``3d convolutional neural networks for human
  action recognition,'' \emph{IEEE transactions on pattern analysis and machine
  intelligence}, vol.~35, no.~1, pp. 221--231, 2013.

\bibitem{schuster1997bidirectional}
M.~Schuster and K.~K. Paliwal, ``Bidirectional recurrent neural networks,''
  \emph{IEEE Transactions on Signal Processing}, vol.~45, no.~11, pp.
  2673--2681, 1997.

\bibitem{nair2010rectified}
V.~Nair and G.~E. Hinton, ``Rectified linear units improve restricted
  {B}oltzmann machines,'' in \emph{Proceedings of the 27th International
  Conference on Machine Learning (ICML-10)}, 2010, pp. 807--814.

\bibitem{abadi2016tensorflow}
M.~Abadi, A.~Agarwal, P.~Barham, E.~Brevdo, Z.~Chen, C.~Citro, G.~S. Corrado,
  A.~Davis, J.~Dean, M.~Devin \emph{et~al.}, ``Tensorflow: Large-scale machine
  learning on heterogeneous distributed systems,'' \emph{arXiv preprint
  arXiv:1603.04467}, 2016.

\bibitem{tflearn2016}
A.~Damien \emph{et~al.}, ``Tflearn,'' \url{https://github.com/tflearn/tflearn},
  2016.

\bibitem{tieleman2012lecture}
T.~Tieleman and G.~Hinton, ``Lecture 6.5-rmsprop: Divide the gradient by a
  running average of its recent magnitude,'' \emph{COURSERA: Neural Networks
  for Machine Learning}, vol.~4, no.~2, 2012.

\bibitem{jia2014caffe}
Y.~Jia, E.~Shelhamer, J.~Donahue, S.~Karayev, J.~Long, R.~Girshick,
  S.~Guadarrama, and T.~Darrell, ``Caffe: Convolutional architecture for fast
  feature embedding,'' in \emph{Proceedings of the 22nd ACM international
  conference on Multimedia}.\hskip 1em plus 0.5em minus 0.4em\relax ACM, 2014,
  pp. 675--678.

\bibitem{scikit-learn}
F.~Pedregosa, G.~Varoquaux, A.~Gramfort, V.~Michel, B.~Thirion, O.~Grisel,
  M.~Blondel, P.~Prettenhofer, R.~Weiss, V.~Dubourg, J.~Vanderplas, A.~Passos,
  D.~Cournapeau, M.~Brucher, M.~Perrot, and E.~Duchesnay, ``Scikit-learn:
  Machine learning in {P}ython,'' \emph{Journal of Machine Learning Research},
  vol.~12, pp. 2825--2830, 2011.

\bibitem{chung2014empirical}
J.~Chung, C.~Gulcehre, K.~Cho, and Y.~Bengio, ``Empirical evaluation of gated
  recurrent neural networks on sequence modeling,'' \emph{arXiv preprint
  arXiv:1412.3555}, 2014.

\bibitem{cybenko1989approximation}
G.~Cybenko, ``Approximation by superpositions of a sigmoidal function,''
  \emph{Mathematics of control, signals and systems}, vol.~2, no.~4, pp.
  303--314, 1989.

\bibitem{shotton2013real}
J.~Shotton, T.~Sharp, A.~Kipman, A.~Fitzgibbon, M.~Finocchio, A.~Blake,
  M.~Cook, and R.~Moore, ``Real-time human pose recognition in parts from
  single depth images,'' \emph{Communications of the ACM}, vol.~56, no.~1, pp.
  116--124, 2013.

\bibitem{mallick2014characterizations}
T.~Mallick, P.~P. Das, and A.~K. Majumdar, ``Characterizations of noise in
  kinect depth images: a review,'' \emph{IEEE Sensors Journal}, vol.~14, no.~6,
  pp. 1731--1740, 2014.

\end{thebibliography}
\bibliographystyle{IEEEtran}

\end{document}